\documentclass[conference]{IEEEtran}
\IEEEoverridecommandlockouts
\usepackage{graphicx}
\usepackage{subfigure}
\usepackage{graphics,color}
\usepackage{booktabs} 
\usepackage{amssymb,amsmath}
\usepackage[symbol]{footmisc}
\usepackage{algorithm}
\usepackage{algorithmic}
\usepackage{physics}
\usepackage{float}
\usepackage{multirow}
\usepackage{paralist}
\usepackage{hyperref}
\usepackage[backend=bibtex,firstinits=true, maxnames=5]{biblatex}
\usepackage{color}
\usepackage{xcolor}
\usepackage[english]{babel}
\usepackage{amsthm}
\theoremstyle{definition}
\newtheorem{definition}{Definition}
\newtheorem{theorem}{Theorem}

\definecolor{darkgray}{rgb}{0.66, 0.66, 0.66}
\definecolor{RoseQuartzBg}{HTML}{F7CAC9}
\definecolor{RoseQuartz}{HTML}{F5A798}
\definecolor{Serenity}{HTML}{92A8D1}
\definecolor{OrangeRed}{rgb}{1.0, 0.27, 0.0}
\definecolor{Red}{rgb}{1.0, 0.0, 0.0}
\definecolor{Turquoise}{HTML}{0F4C81}

\NewDocumentCommand{\zhou}{ mO{} }{\textcolor{red}{\textsuperscript{\textit{Zhou}}\textsf{\textbf{\small[#1]}}}}
\NewDocumentCommand{\yuzhen}{ mO{} }{\textcolor{blue}{\textsuperscript{\textit{yuzhen}}\textsf{\textbf{\small[#1]}}}}
\NewDocumentCommand{\jianhui}{ mO{} }{\textcolor{brown}{\textsuperscript{\textit{jianhui}}\textsf{\textbf{\small[#1]}}}}
\def\BibTeX{{\rm B\kern-.05em{\sc i\kern-.025em b}\kern-.08em
    T\kern-.1667em\lower.7ex\hbox{E}\kern-.125emX}}
    
\usepackage[utf8]{inputenc} 
\usepackage[T1]{fontenc}    
\usepackage{hyperref}       
\usepackage{url}            
\usepackage{booktabs}       
\usepackage{amsfonts}       
\usepackage{nicefrac}       
\usepackage{microtype}      
\usepackage{xcolor}         
\usepackage{xspace}
\usepackage{subfigure}

\newcommand{\toolName}{\textsc{TransNet}\xspace}    
\renewcommand{\thefootnote}{\fnsymbol{footnote}}

\usepackage{expl3}
\ExplSyntaxOn
\newcommand\latinabbrev[1]{
  \peek_meaning:NTF . {
    #1\@}%
  { \peek_catcode:NTF a {
      #1.\@ }%
    {#1.\@}}}
\ExplSyntaxOff

\def\eg{\latinabbrev{e.g}}

\def\ie{\latinabbrev{i.e}}
\begin{document}

\title{Augmenting Knowledge Transfer across Graphs}

\author{
\IEEEauthorblockN{ Yuzhen Mao\textsuperscript{\textsection}}
\IEEEauthorblockA{
\textit{Simon Fraser University}\\
Greater Vancouver, BC, Canada \\
yuzhenm@sfu.ca}
\and
\IEEEauthorblockN{ Jianhui Sun}
\IEEEauthorblockA{
\textit{University of Virginia}\\
Charlottesville, VA, USA \\
js9gu@virginia.edu}
\and
\IEEEauthorblockN{ Dawei Zhou}
\IEEEauthorblockA{
\textit{Virginia Tech}\\
Blacksburg, VA, USA \\
zhoud@vt.edu}
}

\maketitle
\begingroup\renewcommand\thefootnote{\textsection}
\footnotetext{This work is done as an undergraduate research assistant in Virginia Tech.}

\begin{abstract}
Given a \emph{resource-rich} source graph and a \emph{resource-scarce} target graph, how can we effectively transfer knowledge across graphs and ensure a good generalization performance?
In many high-impact domains (e.g., brain networks and molecular graphs), collecting and annotating data is prohibitively expensive and time-consuming, which makes domain adaptation an attractive option to alleviate the label scarcity issue.
In light of this, the state-of-the-art methods focus on deriving domain-invariant graph representation that minimizes the domain discrepancy. 
However, it has recently been shown that a small domain discrepancy loss may not always guarantee a good generalization performance, especially in the presence of disparate graph structures and label distribution shifts. 
In this paper, we present \toolName, a generic learning framework for augmenting knowledge transfer across graphs. In particular, we introduce a novel notion named trinity signal that can naturally formulate various graph signals at different granularity (e.g., node attributes, edges, and subgraphs). With that, we further propose a domain unification module together with a trinity-signal mixup scheme to jointly minimize the domain discrepancy and augment the knowledge transfer across graphs. 
Finally, comprehensive empirical results show that \toolName\ outperforms all existing approaches on seven benchmark datasets by a significant margin.
\end{abstract}


\section{Introduction}
Graph provides a pivotal data structure and a fundamental abstraction for modeling many complex systems, ranging from social science to material science, from financial fraud detection to traffic prediction and many more. The success of convolutional neural networks (CNNs)~\cite{kipf2016semi} for grid data has inspired the recent development of graph neural networks (GNNs), which have achieved superior performance on a variety of graph mining tasks such as node classification, link prediction, subgraph matching, and network alignment. Despite the remarkable success, the performance of GNNs is largely attributed to the abundant and high-quality training data. However, in many high-impact domains (e.g., brain networks and molecular graphs), there exist only scarce labels as the data annotation process is prohibitively expensive and time-consuming. Therefore, a fundamental problem is how to transfer knowledge from the resource-rich source graph to the resource-scarce target graph and ensure a good generalization performance. 

Domain adaptation is an attractive solution to tackle this problem, which has received a surge of attention~\cite{hu2020graph,wu2020unsupervised} in the graph mining community. The general philosophy is to learn \emph{domain-invariant representations} that do not only achieve satisfactory source domain performances, but also generalize well to the label-scarce target domain.
Abundant algorithms~\cite{ganin2016domain} and statistical guarantees~\cite{ben2010theory, blitzer2007learning} have been proposed specifically for the independent and identically distributed (i.i.d.) data. However, how to generalize these algorithms and theoretical results to the graph-structured data
(i.e., instances are apparently non-iid due to the interconnecting nodes and edges) with heterogeneous graph signals (e.g., node, edges, motifs) is under-explored. Moreover, recent studies~\cite{wu2019domain} have shown that domain-invariant representation may not be able to guarantee a good generalization performance, especially in the presence of disparate graph structures and label distribution shifts, which motivates us to propose novel approach with rigorous guarantees to improve the generalization performance of GNNs across graphs.

Towards this goal, we identify the following two challenges: 
\emph{C1. Graph Discrepancy}: how to eliminate negative transfer when the source graph and target graph exhibit disparate structures and feature spaces?
\emph{ C2. Signal Heterogeneity}:  how to effectively characterize and leverage graph signals which are heterogeneous (e.g., node, edges, motifs) in both source and target graphs to improve the generalization performance?

In this paper, we propose a generic learning framework named \toolName\ for augmenting knowledge transfer across graphs and show that our proposed approach achieves superior performances universally on all backbone GNNs.
The main idea behind our method is a principled way to unify the heterogeneous signals on disparate graphs.
To address C1, we develop bi-level gradient reversal layers that learn invariant representations to unify the structure and feature space of the source and target graphs. To address C2, we firstly introduce a novel notion named trinity signal that can naturally formulate various graph signals (e.g., node attributes, edges, and subgraphs). That is to say, we can transform heterogeneous graph signals into a unified format. Building upon this, we propose a data augmentation scheme that automatically conducts interpolation and mixup upon trinity signals to regularize the backbone GNNs with a smooth decision boundary. In general, our contributions are summarized as follows.
\begin{itemize}
    \item \textbf{Problem.} We formalize the \emph{graph signal domain adaptation} problem and identify multiple unique challenges inspired by the real applications. 
    \item \textbf{Algorithm.} We propose a novel method named \toolName\ that (1) unifies the heterogeneous graph signals and dissipate feature spaces and (2) automatically augments the knowledge transfer via trinity-signal mixup. 
    \item \textbf{Evaluation.} We systematically evaluate the performance of \toolName\ on seven real graphs by comparing them with eleven baseline models, which verifies the efficacy of \toolName.  
    We find that \toolName\ largely alleviates the negative transfer issue and leads up to 9.45\% precision improvement over the state-of-the-art methods. 
    \item \textbf{Reproducibility. }We publish our data and code at~\url{https://github.com/yuzhenmao/TransNet}
\end{itemize}

The rest of our paper is structured as follows. The problem definition is introduced in Section II, followed by the discussion of \toolName in Section III. Experimental results are reported in Section IV. In Section V, we review the existing literature before we conclude the paper in Section VI.

\section{Problem Definition}

In the setting of domain adaptation across graphs, we denote the source graph $\mathcal{G}_s$ and the target graph $\mathcal{G}_t$ in the form of triplets, i.e.\  $\mathcal{G}_s = (\mathcal{V}_s, \mathcal{E}_s, \mathbf{X}_s)$ and $\mathcal{G}_t = (\mathcal{V}_t, \mathcal{E}_t, \mathbf{X}_t)$, where $\mathcal{V}_s$ ($\mathcal{V}_t$) represents the set of nodes, $\mathcal{E}_s$ ($\mathcal{E}_t$) represents the set of edges, and $\mathbf{X}_s$ ($\mathbf{X}_t$) represents the node features in $\mathcal{G}_s$ ($\mathcal{G}_t$). Moreover, we denote the adjacency matrices of $\mathcal{G}_s$ and $\mathcal{G}_t$ as $\mathbf{A}_s$ and $\mathbf{A}_t$ correspondingly. 
The goal of this paper is to translate the relevant and complementary information from the source graph to the target graph, by addressing graph discrepancy and signal heterogeneity. 

\begin{figure}[htp]
\centering
\includegraphics[width=1\linewidth]{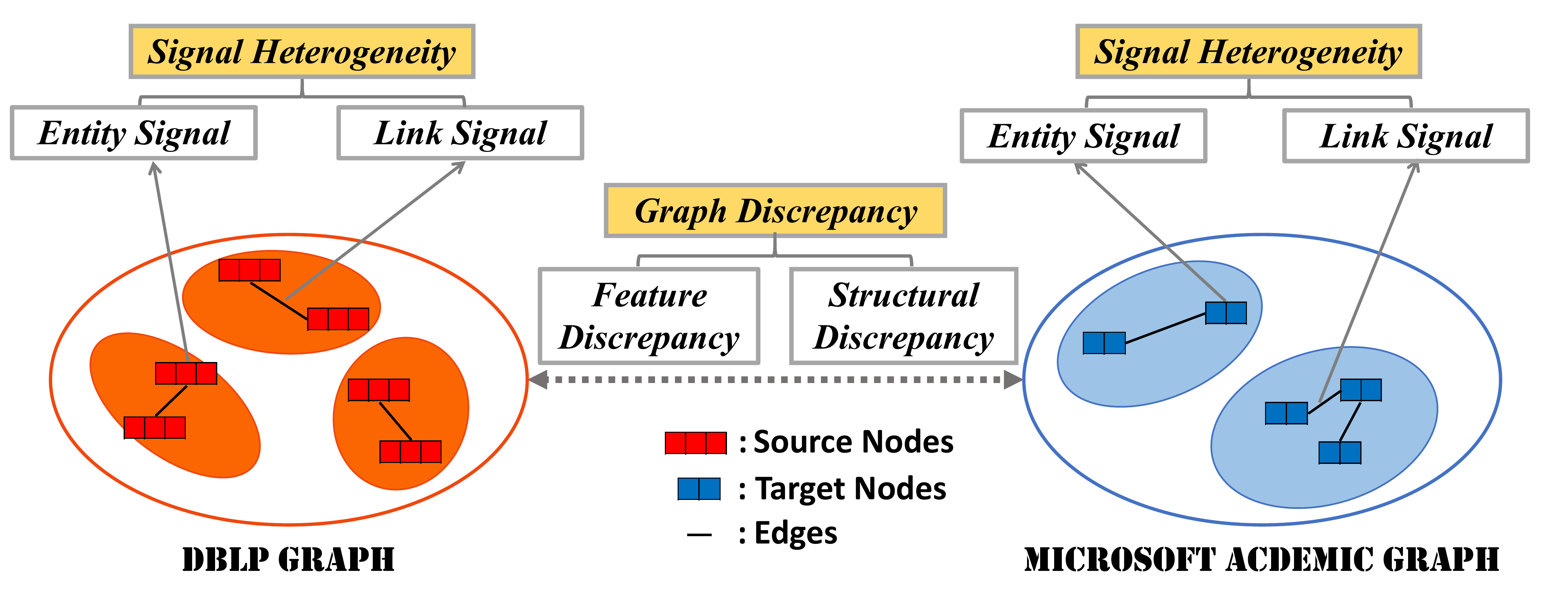}
\caption{An illustrative example of domain adaptation across DBLP graph and Microsoft Academic Graph.}
\label{fig:illustration}
\end{figure}

\noindent\textbf{Problem Definition}
We consider transferring knowledge learned from the source domain(s) to a target domain with limited labels. Fig~\ref{fig:illustration} presents an illustrative example, which visualizes knowledge transfer from the DBLP Graph ($\mathcal{G}_s$) to the Microsoft Academic Graph ($\mathcal{G}_t$). Here, both source and target domain data could be modeled as graphs. 
As shown in Fig~\ref{fig:illustration}, there are two obstacles, including graph discrepancy and signal heterogeneity during graph domain adaptation. On the one hand, real-world graphs are complex and composed of heterogeneous signals, including entity signals (\eg, nodes, subgraphs) and the corresponding link signals between them. On the other hand, graphs across different domains naturally exhibit disparate distribution in feature representations (\eg, different feature dimensions in $\mathcal{G}_s$ and $\mathcal{G}_t$) and structural organizations (\eg, three clusters in $\mathcal{G}_s$ while two clusters in $\mathcal{G}_t$). Given that, we formally define our problem as follows:

\noindent\textbf{Problem 1.}  Knowledge Transfer across Graphs.\\
\noindent\textbf{Given:} The source graph $\mathcal{G}_s = (\mathcal{V}_s, \mathcal{E}_s, \mathbf{X}_s)$ with rich node labels $\mathcal{Y}_s$, the target graph $\mathcal{G}_t = (\mathcal{V}_t, \mathcal{E}_t, \mathbf{X}_t)$ with few-shot node labels $\mathcal{\tilde{Y}}_t \in \mathcal{Y}_t$.\\
\noindent\textbf{Find:} Accurate predictions $\hat{\mathcal{Y}_t}$ of unlabeled examples in the target graph $\mathcal{G}_t$.

\begin{figure}[t]
\centering
\includegraphics[width=1\linewidth]{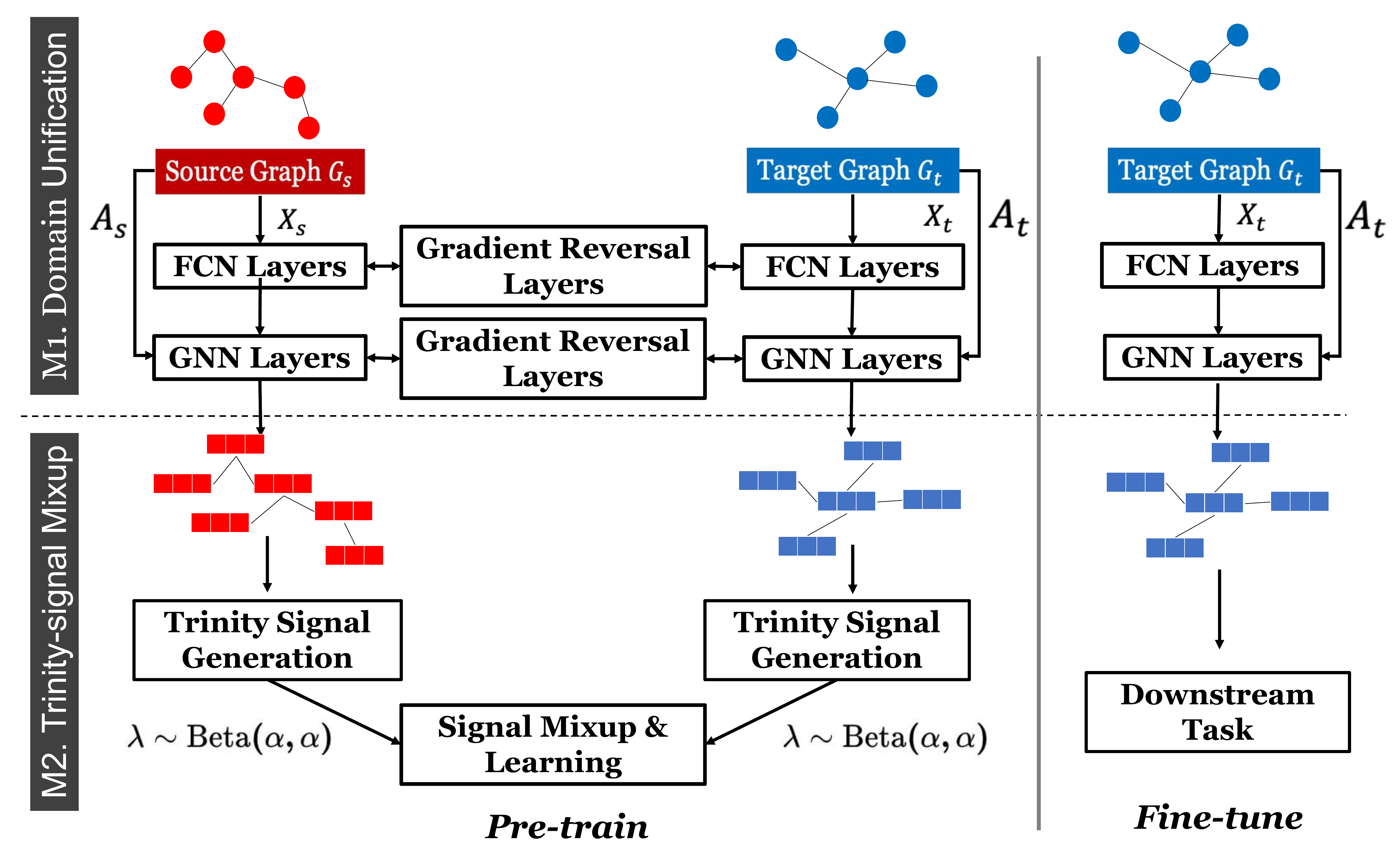}
\caption{The proposed \toolName framework.}
\label{fig:framework}
\end{figure}
\section{Methodology}

We first review graph pre-training strategies and a theoretical model for domain adaptation before diving into our model.

\textbf{Graph Pre-training.}
Graph pre-training strategies~\cite{hu2019strategies, hu2020gpt, jin2020self,zhou2022mentorgnn} provide a powerful tool to parameterize GNNs without label information by predicting easily-accessible graph signals (e.g., node/edge features, context information~\cite{hu2019strategies}, distance2clusters~\cite{jin2020self}) extracted from the input graph. In general, the learning objective of existing graph pre-training strategies can be formulated as follows
\begin{equation}\label{Eq: pretrain}
\text{argmax}_{\mathbf{\theta}} \mathbb{E}_{\mathbf{s}\in \mathcal{G}} \log h_\mathbf{\theta}(\mathbf{s}|\hat{\mathcal{G}},\mathbf{\theta})
\end{equation}
where $\mathcal{G}$ is the input graph, $\hat{\mathcal{G}}$ is the corrupted graph with some masked graph signals $\mathbf{s}$, $h(\cdot)$ is a GNN model with hidden parameters $\theta$. Open research questions lie in how to effectively pre-train GNNs in the presence of heterogeneous graph signals. 

\textbf{Domain Adaptation.}
A domain consists of a distribution $\mathcal{D}$ on space $\mathcal{X}$ and a labeling function $f:\mathcal{X} \rightarrow[0,1].$ Given two domains, a source domain $\left\langle\mathcal{D}_{s}, f_{s}\right\rangle$ and a target domain $\left\langle\mathcal{D}_{t}, f_{t}\right\rangle$, as well as a hypothesis $ h: \mathcal{X} \rightarrow\{0,1\}$, we define the risk of the hypothesis $h(\cdot)$ w.r.t. a true labeling function $f(\cdot)$ under distribution $\mathcal{D}$ as $\epsilon(h)=\mathrm{E}_{\mathbf{x} \sim \mathcal{D}}[|h(\mathbf{x})-f(\mathbf{x})|]$. As a common notion, the empirical risk of a function $h(\cdot)$ on the source domain is defined as $\hat{\epsilon}_{s}(h)$. Similarly, for the target domain, we use the parallel notation
$\epsilon_{t}(h)$, and $\hat{\epsilon}_{t}(h)$. In ~\cite{ben2010theory} and ~\cite{blitzer2007learning}, the generalization bound on the target risk in terms of the empirical source risk and the discrepancy between the source and target domains is derived as follows

\begin{theorem}[\cite{blitzer2007learning}]
With probability at least $1-\delta$, for every $h \in \mathcal{H}$,

\begin{equation}
\begin{aligned}
    \varepsilon_{t}(h) \leq & \widehat{\varepsilon}_{s}(h)+\frac{1}{2} d_{\mathcal{H} \Delta \mathcal{H}}\left(\widehat{\mathcal{D}}_{s}, \widehat{\mathcal{D}}_{t}\right)+\lambda\\
    & + O\left(\sqrt{\frac{d \log (m / d)+\log (1 / \delta)}{m}}\right)
\end{aligned}
\end{equation}
where $\widehat{\mathcal{D}}_{s} ( \widehat{\mathcal{D}}_{t})$ denotes the empirical distribution induced by $m$ samples drawn from ${\mathcal{D}}_{s} ({\mathcal{D}}_{t})$; $\mathcal{H}$ denotes a hypothesis class; ${d}_{\mathcal{H} \Delta \mathcal{H}}$ denotes the distance on $(\widehat{\mathcal{D}}_{s}, \widehat{\mathcal{D}}_{t})$ induced by the symmetric difference hypothesis space; $\lambda$ denotes the combined risk of the optimal hypothesis; and the last term is a constant which does not depend on any particular $h(\cdot)$.
\end{theorem}

\subsection{A Generic Learning Framework}
In the rest of this section, we propose \toolName, a generic learning framework that aims to augment knowledge transfer from the source graph to the target graph. 
An overview of \toolName is presented in Fig~\ref{fig:framework}, which consists of two major modules: \emph{M1. Domain Unification} and \emph{M2. Trinity-signal Mixup}. These two modules are designed to address C1 and C2, correspondingly. In particular, to address the graph discrepancy challenge (C1), M1 automatically unifies the disparate structure and feature distributions of $\mathcal{G}_s$ and $\mathcal{G}_t$ into a domain-invariant hidden space; to address the signal heterogeneity challenge (C2), M2 further unifies the formats of heterogeneous graph signals and conducts manifold mixup~\cite{verma2019manifold} operation to achieve a smooth decision boundary. We will further rationale the significance of these two modules with ablation studies (Section~\ref{sec: ablation}). 
In the following subsections, we dive into the two modules of \toolName in detail. 

\textbf{M1. Domain Unification. }
Learning invariant representations is crucial for efficient knowledge transfer. One of the standard adversarial approaches is minimizing the distribution discrepancy between domains by Gradient Reversal Layer (GRL)~\cite{ganin2016domain}. However, domain adaptation on graph-structured data naturally exhibits the bi-level discrepancy (\ie, feature discrepancy and structural discrepancy), which is illustrated in Figure~\ref{fig:illustration}. Different from the previous methods~\cite{shen2020adversarial, shen2020network, dai2022graph}, here we propose a bi-level GRL scheme (shown in M1 of Figure~\ref{fig:framework}) to unify the structure and feature space discrepancy of different domains. 
Firstly, given raw nodes representations $\mathbf{x}_s \in \mathbf{X}_s$ and $\mathbf{x}_t \in \mathbf{X}_t$, we develop domain-specific feature encoder functions that transform $\mathbf{x}_s$ and $\mathbf{x}_t$ to a small domain-invariant hidden space. To eliminate the feature discrepancy, we implement the feature encoder function via Multi-Layer Perceptron (MLP) regularized by GRL. 
Next, by obtaining the unified node feature representations, we feed them forward to a shared Graph Neural Network (GNN) for extracting domain-invariant structural information, which is also regularized by GRL. 
By regularizing feature discrepancy and structural discrepancy via M1, we are able to encode $\mathbf{x}_s$ and $\mathbf{x}_t$ into a domain-invariant space. 
In particular, we formulate the loss function $\mathcal{L}_{\text {domain}}$ of M1 as follow
\begin{equation}
\begin{aligned}
\mathcal{L}_{\text {domain}} =& {\tt Unif}_f + {\tt Unif}_s \\
=& \underbrace{{\tt GRL}({\tt MLP} (\mathbf{x}_s), {\tt MLP} (\mathbf{x}_t))}_{{\tt Unif}_f\text{: feature discrepancy loss}} \\
&\underbrace{+ {\tt GRL}( {\tt GNN} ( {\tt MLP} (\mathbf{x}_s), \mathbf{A}_s), {\tt GNN} ( {\tt MLP} (\mathbf{x}_t), \mathbf{A}_t))}_{{\tt Unif}_s\text{: structural discrepancy loss}}
\end{aligned}
\end{equation}
where ${\tt Unif}_f$ denotes the feature discrepancy loss, ${\tt Unif}_s$ denotes the structure discrepancy loss. Without M1, downstream trinity-signal mixup module would potentially blend in unnecessarily redundant signals and thus result in negative transfer~\cite{ganin2016domain}. In general, M1  disentangles the domain-specific information by utilizing bi-level GRL and only keeps the domain invariant information, which paves the way for trinity-signal mixup in M2.

\textbf{M2. Trinity-signal Mixup.} 
As graph-structured data is complex and hierarchical, it naturally exhibits heterogeneous signals. To utilize the information encoded in different signals, existing graph pre-training and domain adaptation approaches treat each signal separately, \eg, ~\cite{hu2019strategies} and ~\cite{hu2020gpt} design different pre-train tasks for different signals, while~\cite{wu2020unsupervised} applies an attention scheme to capture the significances of different signals. This could lead to high learning complexity and limit the usage of several useful techniques (\eg, mixup~\cite{zhang2017mixup} and data poisoning). Here, inspiring from multi-label learning, we propose a generic data structure named trinity signal to unify the representation of heterogeneous graph signals with multi-labels as follows

\begin{definition}[Trinity Signal]~\label{def}
Given a pair of connected signals $\{\mathbf{s}_i, \mathbf{s}_j\}$ in graph $\mathcal{G}$ together with their representations $\{\mathbf{e}_i, \mathbf{e}_j\}$, the corresponding labels $\{ y_i, y_j\}$ and connection property $p_{ij}$, the trinity signal representation of $\{\mathbf{s}_i, \mathbf{s}_j\}$ is defined as: $\mathbf{t}_{ij} = {\tt MLP}(\left[\mathbf{e}_{i}, \mathbf{e}_{j}\right])$ with multi-labels $\mathbf{y}_{ij} = \{y_{i}, y_{j}, p_{ij} \}$, where $\left[ \cdot \right]$ denotes the concatenation operation.
\end{definition}

In practice, the trinity signals can be generalized to various graph signals. For instance, when $\mathbf{s}_{i}$ and $\mathbf{s}_{j}$ represents a pair of nodes, then $\mathbf{e}_{i}$ ($\mathbf{e}_{j}$) denotes the node representation, $y_{i}$ ($y_{j}$) denotes the node label, $p_{ij}$ denotes the weight or proximity score between $\mathbf{s}_{i}$ and $\mathbf{s}_{j}$ (\eg, edge existence and personalized PageRank); when $\mathbf{s}_{i}$ ($\mathbf{s}_{j}$) denotes a (sub)graph~\cite{hu2019strategies}, similarly, $\mathbf{e}_{i}$ ($\mathbf{e}_{j}$) denotes a (sub)graph representation, $y_{i}$ ($y_{j}$) denotes a (sub)graph label, $p_{ij}$ denotes the (sub)graph distance between $\mathbf{s}_{i}$ and $\mathbf{s}_{j}$ (\eg, graph similarity or graph edit distance). In general, trinity signals simultaneously encode entity signals (\eg, nodes, subgraphs) and the corresponding link signals in a principled way. 

However, after unifying heterogeneous graph signals, discreteness and  non-differentiability still exist in the generated trinity signals, which leads to sub-optimal performance of the model~\cite{liu2020unified}.
Mixup~\cite{zhang2017mixup}, a widely adopted data augmentation technique, is a potential approach which has been shown to improve both generalizability and robustness in various domains~\cite{zhang2020does}. 
Motivated by this, we propose a novel graph mixup strategy named trinity-signal mixup that could be conducted upon the trinity graph signals. 
Formally, given two trinity signals $\mathbf{t}$ and $ \mathbf{t^{\prime}}$ with labels $\mathbf{y} = \{y_{1}, y_{2}, p\}$ and $\mathbf{y}^{\prime} = \{y_{1}^{\prime}, y_{2}^{\prime}, p^{\prime}\}$ respectively, we firstly map the trinity signals to a latent space by one linear fully connected layer. Then, a mixup function ${\tt Mixup}_{\lambda}\left(\mathbf{t}, \mathbf{t^{\prime}} \right)$ generates a new interpolated trinity signal $\tilde{\mathbf{t}}$, where $\lambda \sim \operatorname{Beta}(\alpha, \alpha)$, for $\alpha \in(0, \infty)$~\cite{zhang2017mixup}:
\begin{equation}
\label{eq: five}
\tilde{\mathbf{t}} = {\tt Mixup}_{\lambda}\left(\mathbf{t}, \mathbf{t^{\prime}} \right) = \lambda * \mathbf{t} + (1-\lambda) * \mathbf{t^{\prime}} 
\end{equation}
with labels defined $\tilde{\mathbf{y}}$  as:
\begin{equation}
\label{eq: six}
\begin{aligned}
\tilde{\mathbf{y}} = {\tt Mixup}_{\lambda}\left(\mathbf{y}, \mathbf{y}^{\prime}\right) = \{\lambda * y_{1} + (1-\lambda) * y_{1}^{\prime}, \\ \lambda * y_{2} + (1-\lambda) *  y_{2}^{\prime}, \\ \lambda * p + (1-\lambda) * p^{\prime}\}
\end{aligned}
\end{equation}

We also train a multi-label classifier $g(\cdot)$ which outputs the label of trinity signals in $\hat{\mathbf{y}}$:
\begin{equation}
\hat{\mathbf{y}}  = \{\hat{y_{1}}, \hat{y_{2}}, \hat{p}\} = g\left({\tt Mixup}_{\lambda}\left(\mathbf{t}, \mathbf{t^{\prime}}\right)\right)
\end{equation}

We define the loss function of trinity-signal mixup as follows
\begin{align}\label{eq: four}
    \mathcal{L}_{signal}\left(\mathcal{D}, \alpha\right)= & \underset{(\mathbf{t}, \mathbf{y}) \sim \mathcal{D}}{\mathbb{E}} \underset{\left(\mathbf{t}^{\prime}, \mathbf{y}^{\prime}\right) \sim \mathcal{D}}{\mathbb{E}} \underset{\lambda \sim \operatorname{Beta}(\alpha, \alpha)}{\mathbb{E}} \\
    & \quad \ell\left(g\left({\tt Mixup}_{\lambda}\left(\mathbf{t}, \mathbf{t}^{\prime}\right)\right), {\tt Mixup}_{\lambda}\left(\mathbf{y}, \mathbf{y}^{\prime}\right)\right) \nonumber
\end{align}
where $\mathcal{D}$ is a specific data distribution, $(\mathbf{t}, \mathbf{y})$ and $(\mathbf{t}^{\prime}, \mathbf{y}^{\prime})$ is a pair of labeled examples sampled from distribution $\mathcal{D}$, $\ell$ is a composite loss function including cross-entropy loss for node classification and mean squared loss for distance regression.
In general, trinity signals provide high flexibility for the end users to handle various graph signals at different granularities (\eg, node-level, edge-level, subgraph-level).

\subsection{Algorithm}
The overall objective function is defined as follows

\begin{equation}
\label{eq: all}
\mathcal{L}_{total}=\mathcal{L}_{\text {domain}} + \gamma*\mathcal{L}_{\text {signal}}
\end{equation}
where $\mathcal{L}_{\text {domain}}$ denotes the bi-level GRL loss, $\mathcal{L}_{\text {signal}}$ denotes the trinity-signal loss, and $\gamma$ is the hyper-parameter that balances the contributions of the two terms.

The procedure for \toolName training is presented in Algorithm~\ref{alg:one}, with Adam as the optimizer. Given the source graph $\mathcal{G}_s = (\mathcal{V}_s, \mathcal{E}_s, \mathbf{X}_s)$ with rich labels $\mathcal{Y}_s$; the target graph $\mathcal{G}_t = (\mathcal{V}_t, \mathcal{E}_t, \mathbf{X}_t)$ with limited labels $\mathcal{\tilde{Y}}_t \in \mathcal{Y}_t$, we hope to learn a model predicting the node labels of the target graphs. 

\begin{algorithm}[t]
\caption{The \toolName\ Learning Framework}
\label{alg:one}
\begin{algorithmic}[1]
\REQUIRE ~~\\
    (i) a source graph $\mathcal{G}_s = (\mathcal{V}_s, \mathcal{E}_s, \mathbf{X}_s)$ with rich labels $\mathcal{Y}_s$; (ii) a target graph $\mathcal{G}_t = (\mathcal{V}_t, \mathcal{E}_t, \mathbf{X}_t)$ with few-shot labels $ \mathcal{\tilde{Y}}_t$; (iii) parameter $k$.
\ENSURE ~~\\
    Predictions $\hat{\mathcal{Y}_t}$ of unlabeled examples in $\mathcal{G}_t$\\
    \STATE Initialize the domain unification model, the trinity-signal classifier $g(\cdot)$, and the classifier $h(\cdot)$ for the downstream task in $\mathcal{G}_t$.
    \WHILE{not convergent}
        \STATE Compute domain-invariant representations of both $\mathcal{G}_s$ and $\mathcal{G}_t$ via domain unification.
        \STATE Generate $k$ trinity signals and apply manifold mixup based on Eq.~\ref{eq: five}\&\ref{eq: six}.
        \STATE Update the hidden parameters of the domain unification model and the trinity-signal classifier $g(\cdot)$ by minimizing the overall loss function in Eq.~\ref{eq: all}. 
    \ENDWHILE
    \WHILE{not convergent}
        \STATE Fine-tune ${\tt MLP}$ of the target domain, the ${\tt GNN}$ and the classifier $h(\cdot)$ for the downstream task.
    \ENDWHILE
\end{algorithmic}
\end{algorithm}


\section{Experiment}\label{exp}
In this section, we demonstrate the performance of our proposed model \toolName on seven benchmark datasets by comparing with eleven state-of-the-art baselines. 

\subsection{Experiment Setup}
\textbf{Datasets:}
We evaluate \toolName on seven real-world undirected graphs, including five paper citation graphs: Microsoft Academic Graph~\cite{shen2020adversarial}, DBLPv7~\cite{shen2020adversarial}, DBLPv8~\cite{wu2020unsupervised}, ACMv9\_{1}~\cite{wu2020unsupervised}, ACMv9\_{2}~\cite{shen2020adversarial}, where nodes represent papers, edges represent a citation relation between two linked nodes; and two co-purchase graphs~\cite{shchur2018pitfalls}: Amazon Computers, Amazon Photo, where nodes represent goods, edges represent that two linked goods are frequently bought together. All these seven datasets use bag-of-words encoded features, and each node is associated with one label only. In this paper, we use A1, D1, A2, M2, D2, Comp, Photo to denote ACMv9\_{1}, DBLPv8, ACMv9\_{2}, Microsoft Academic Graph, DBLPv7, Amazon Computers, Amazon Photo, respectively.

\textbf{Comparison Baselines:} We compare \toolName with five GNNs, two graph pre-train methods, and four graph transfer learning methods.

\noindent\underline{GNNs}: \textbf{GCN}~\cite{kipf2016semi}, \textbf{GAT}~\cite{velivckovic2017graph}, \textbf{GIN}~\cite{xu2018powerful}, \textbf{GraphSAGE}~\cite{hamilton2017inductive} are four standard graph representation benchmark architectures. \textbf{GraphMix}~\cite{verma2019graphmix2} is one of the most popular graph mixup model.\\
\underline{Graph Pre-train}: \textbf{GPT}~\cite{hu2020gpt} pre-trains a GNN by introducing a self-supervised attributed graph generation task. \textbf{SelfTask}~\cite{jin2020self} builds advanced pretext tasks to pre-train the GNN. \\
\underline{Transfer Learning on Graphs}: \textbf{GPA}~\cite{hu2020graph} is a transferable active learning model. \textbf{DANN}~\cite{ganin2016domain} is a classical domain adaptation method with GRL. In our experiment, we use GCN as its feature extractor. \textbf{UDAGCN}~\cite{wu2020unsupervised} and \textbf{ACDNE}~\cite{shen2020adversarial} are two domain adaptation methods for graph structured data. 

For a fair comparison, all baselines contain two GNN hidden layers with $d_{1}=64$ and $d_{2}=32$ for the first and second layers, respectively. The output dimension of GNN is 16. We conduct experiments with only five labeled samples in each class of the target dataset and test based on the rest unlabeled nodes. For UDAGCN and ACDNE having the constraints of shared input features, we follow the instruction from the original papers~\cite{wu2020unsupervised, shen2020adversarial} to build a union set for input features between the source and target domains by setting zeros for unshared features. For classical GNNs (GCN, GAT, GIN, GraphSage), we directly train each model on the target domain for 2000 epochs. For domain adaptation models (DANN, UDAGCN, ACDNE), after training from the source datasets, they are fine-tuned on the target datasets for 1000 epochs. 

For \toolName, it is firstly pre-trained on the source dataset for 2000 epochs; then it is fine-tuned on the target dataset for 800 epochs using limited labeled data in each class. We use Adam optimizer with learning rate 3e-3. $\alpha$ in the beta-distribution of trinity-signal mixup is set to 1.0. The output dimension of $\tt{MLP}$ in domain unification module is set to 100. Precision is used as the evaluation metric. We run the experiments with 100 random seeds. The experiments are performed on a Ubuntu20 machine with 16 3.8GHz AMD Cores and a single 24GB NVIDIA GeForce RTX3090.

\subsection{Effectiveness}\label{sec: ablation}
\textbf{Comparison Results.} We compare \toolName with eleven baseline methods across seven real-world undirected graphs. We show the precision of different methods in Table \ref{tab:results}. In general, we have the following observations: 
(1) Our proposed \toolName consistently outperforms all the baselines on seven datasets, which demonstrates the generalizability and effectiveness of our model. Especially, when adapting knowledge from DBLPv8 to Microsoft Academic Graph with five labeled samples per class, the improvement is more than 10\% comparing with the second best model (DANN).
(2) Classical GNNs have good performance in several datasets including DBLPv7 and Amazon Computers; but in most instances, they have relatively lower precision. For example, in dataset ACMv9\_2, with five labeled samples per class, the best precision is 50.18\% achieved by GNN, which is 5\% lower than GPT and 14\% lower than \toolName. The reason is that these models don't make use of the additional knowledge from the source graph, which leads to relatively worse performance especially when the labeled samples are limited. 
(3) Graph pre-train models sometimes achieve significant improvement: SelfTask and GPA outperform all classical GNNs in dataset ACMv9\_2 and DBLPv7 respectively. But compared with \toolName, they have relatively poor generalization performance since these pre-train models do not consider the graph discrepancy so that they cannot make use of the knowledge from the resource-rich source domains. 
(4) Graph transfer learning models such as DANN and UDAGCN could achieve better performance than classical GNNs and graph pre-train models. Particularly, DANN outperforms all the models except \toolName in datasets ACMv9\_2, Microsoft Academic Graph, and DBLPv7 with both three or five labeled samples per class. However, \toolName could still beat graph transfer learning models in every dataset. For example, in datasets ACMv9\_2, Microsoft Academic Graph, and DBLPv7, \toolName outperforms all listed graph transfer learning models by at least 5\% precision. Comparing with graph transfer learning models, the key advantage of \toolName\ lies in the trinity-signal mixup that could handle signal heterogeneity and reduce the learning complexity simultaneously.

\begin{table*}[htbp]
\caption{Comparison of different methods using 5 labeled samples per class (\% test precision).}
\setlength{\tabcolsep}{3pt}
\centering
\scalebox{0.97}{
\begin{tabular}{ccccccccccccccc}
\cline{1-14}
 Source        & Target              & GCN                    & GAT                    & GIN                    & GraphSAGE      &GraphMix        & GPT-GNN & SelfTask & GPA & DANN & UDAGCN & ACDNE & \multicolumn{1}{|c}{\toolName}       \\ \cline{1-14}
 Photo & Comp        & 67.24                  & 65.81                 & 66.37                  & 71.26        &      42.13    &   62.75     & 63.18  & 60.22    &     71.74  &     73.13    &   24.55    & \multicolumn{1}{|c}{\textbf{76.54}} \\ \cline{1-14}
 Comp  & Photo       & 79.17                  & 71.58                  & 75.32                  & 84.56            &   74.36   &   75.63      & 76.80 &   71.36  &    83.75   &     81.24    &    33.38   & \multicolumn{1}{|c}{\textbf{87.67}} \\ \cline{1-14}
 A1            & \multirow{2}{*}{A2} & \multirow{2}{*}{50.18} & \multirow{2}{*}{46.90} & \multirow{2}{*}{43.56} & \multirow{2}{*}{45.63} & \multirow{2}{*}{48.64}  & \multirow{2}{*}{55.04}    &  \multirow{2}{*}{46.60}   &   52.02  &   55.34    &    39.33     &   33.14    & \multicolumn{1}{|c}{\textbf{64.11}}          \\
 D1            &                     &                        &                        &                        &                 &       &      &     &   51.60  &  56.35     &   38.66      &  31.71     & \multicolumn{1}{|c}{\textbf{65.34}} \\ \cline{1-14}
 A1            & \multirow{2}{*}{M2} & \multirow{2}{*}{59.60} & \multirow{2}{*}{51.86} & \multirow{2}{*}{47.88} & \multirow{2}{*}{53.17} & \multirow{2}{*}{55.67} & \multirow{2}{*}{64.27}     &   \multirow{2}{*}{54.75}  &  62.53   &    65.75   &      45.90   &   43.77    & \multicolumn{1}{|c}{\textbf{73.53}} \\
 D1            &                     &                        &          &              &                        &                        &      &     &   61.63  &  64.63     &      45.20   &   43.11    & \multicolumn{1}{|c}{\textbf{74.20}}          \\ \cline{1-14}
 A1            & \multirow{2}{*}{D2} & \multirow{2}{*}{58.30} & \multirow{2}{*}{53.39} & \multirow{2}{*}{45.07} & \multirow{2}{*}{52.16} & \multirow{2}{*}{51.59} & \multirow{2}{*}{51.84}   &  \multirow{2}{*}{59.05}  & 57.50    &    60.01   &     42.36    &    40.93   & \multicolumn{1}{|c}{\textbf{66.75}} \\
 D1            &                     &                        &                        &                        &              &          &      &     & 56.89    &   61.87    &   42.26      &    40.17   & \multicolumn{1}{|c}{\textbf{67.95}}          \\ \cline{1-14}
  A2            & \multirow{3}{*}{A1} & \multirow{3}{*}{63.36} & \multirow{3}{*}{62.23} & \multirow{3}{*}{46.82} & \multirow{3}{*}{57.70} &  \multirow{3}{*}{60.17} & \multirow{3}{*}{58.53}  &    \multirow{3}{*}{56.73} &  59.05   &    62.22   &    61.23     &   42.64    & \multicolumn{1}{|c}{\textbf{65.99}}          \\
 M2            &                     &                        &               &         &                        &                        &      &     & 56.56    &    61.63   &      59.27   &   44.75   & \multicolumn{1}{|c}{\textbf{64.74}} \\
 D2            &                     &                        &              &          &                        &                        &      &     & 58.31    &     61.58  &      60.21   &    43.10   & \multicolumn{1}{|c}{\textbf{64.46}}          \\ \cline{1-14}
 A2            & \multirow{3}{*}{D1} & \multirow{3}{*}{94.74} & \multirow{3}{*}{97.33} & \multirow{3}{*}{96.81} & \multirow{3}{*}{94.75} & \multirow{3}{*}{91.15} & \multirow{3}{*}{64.64}    &  \multirow{3}{*}{91.67}  &  68.78   &    95.01   &      91.72   &    29.17   & \multicolumn{1}{|c}{\textbf{97.95}}          \\
 M2            &                     &                        &             &           &                        &                        &      &     &  69.99   &    95.10  &      93.87    &   26.70    & \multicolumn{1}{|c}{\textbf{97.71}}          \\
 D2            &                     &                        &          &              &                        &                        &      &     &  71.27   &    95.44   &   93.54      &   33.45    & \multicolumn{1}{|c}{\textbf{97.91}} \\ \cline{1-14}
\end{tabular}
}
\label{tab:results}
\end{table*}

\textbf{Ablation Study.}
Considering that \toolName consists of various components, we set up the experiments to study the effect of different components by removing one component from \toolName at a time. The ablation results are presented in Table \ref{tab:ablation}. From the results, we have several interesting observations. 
(1) Adding node signals could make a huge improvement to label prediction precision. 
(2) Although adding link signals does not help much in the node classification task (which is reasonable since link signals have no direct connection with node signals), it does not reduce the precision either, which means our model could encode those two signals well simultaneously. 
(3) Although removing the target domain label information could still transfer knowledge, adding target domain influence during the pre-training does make knowledge adaptation even better.
(4) Both two domain losses help the model better adapt knowledge from the source to the target domain, which proves the effectiveness of bi-level GRL in alleviating the graph discrepancy.
(5) Trinity-signal Mixup also helps the model to adapt knowledge better by at most 4\% (DBLPv8 $\rightarrow$ Microsoft Academic Graph).

\begin{table*}[htbp]
\caption{Ablation study using 5 labeled samples per class. Mean and standard deviation are reported over fifty random trials.}
\centering
\scalebox{0.97}{
\begin{tabular}{lcccc}
\hline \textbf{Ablation} & A1 $\rightarrow$ M2  & D1 $\rightarrow$ M2 & A1 $\rightarrow$ A2 & D1 $\rightarrow$ A2\\
\hline Without node signals in source \& target domain & $60.06 \pm 5.91$ & $58.14 \pm 6.05$ & $49.20 \pm 4.94$ & $49.10 \pm 5.86$ \\
Without link signals in source \& target domain & $73.51\pm 3.93$ & $73.72 \pm 3.79$ & $63.78 \pm 5.08$ & $65.28 \pm 4.79$\\
Without target domain node and link signals & $70.43 \pm 3.91$ & $70.55 \pm 3.91$ & $60.29 \pm 3.76$ & $59.95 \pm 3.78$\\
Without ${\tt Unif}_f$ & $50.07 \pm 9.49$ & $55.49 \pm 13.51$ & $47.01 \pm 8.89$ & $41.36 \pm 10.01$\\
Without ${\tt Unif}_s$ & $67.80 \pm 4.05$ & $67.50 \pm 3.21$ & $58.69 \pm 4.68$ & $57.46 \pm 4.18$\\
Without ${\tt Unif}_f$ \& ${\tt Unif}_s$ & $61.86 \pm 9.20$ & $54.51 \pm 9.36$ & $49.31 \pm 10.03$ & $49.72 \pm 7.83$\\
Without Trinity-signal Mixup & $70.54 \pm 4.15$ & $70.83 \pm 3.53$ & $62.57 \pm 4.27$ & $61.22 \pm 5.31$\\ \hline
\toolName & $\textbf{73.53} \pm 4.13$ & $\textbf{74.20} \pm 3.64$ & $\textbf{64.11} \pm 4.75$ & $\textbf{65.34} \pm 5.26$ \\
\hline
\end{tabular}
}
\label{tab:ablation}
\end{table*}

\section{Related Work}
\textbf{Pre-Training for Graphs.} 
Graph pre-training generalizes knowledge to downstream tasks by capturing the structural and semantic properties of input graphs. The current graph pre-training strategies can be summarized into two different categories: 1) Using mutual information maximization between different graph structures which are generated from various corruption functions~\cite{velickovic2019deep}; 2) Utilizing feature generation or edge generation by masking~\cite{hu2020gpt}. Besides, \cite{hu2019strategies} pre-trains a graph at the level of both individual nodes and the entire graph. However, these existing methods cannot transfer knowledge from other domains.

\textbf{Domain Adaptation.} Domain adaptation methods provide potential approach to efficiently transfer knowledge from the source graph to the target graph with disparate structures and label distributions. There are majorly three techniques used for realizing the Domain Adaptation algorithm: 1) Divergence based~\cite{long2015learning,zhou2020domain}; 2) Adversarial based~\cite{ganin2016domain}; 3) Reconstruction based~\cite{bousmalis2016domain}. Recent researches which apply domain adaptation techniques to graph dataset~\cite{wu2020unsupervised,shen2020adversarial, shen2020network, dai2022graph} only focus on the setting of shared input feature. To the best of our knowledge, graph domain adaptation based on two
different input spaces and two output label-sets has received little attention in the machine learning community.

\textbf{Mixup for Data Augmentation.}
Mixup and its variants~\cite{zhang2017mixup, verma2019manifold} are  interpolation-based and widely-adopted data augmentation techniques for regularizing neural networks. More recently, mixup is applied to graph dataset. \cite{verma2019graphmix2} proposes to train an auxiliary Fully-Connected Network which uses the node features to implement Manifold Mixup.~\cite{zhao2021graphsmote} aims to train an edge generator through the task of adjacency matrix reconstruction. \cite{wang2021mixup} mixes the receptive field subgraphs for the paired nodes. These previous works ignore the mixup in the link level or need to use additional networks , which is far less elegant, efficient and accurate.

\section{Conclusion}
In this paper, we present \toolName, a generic learning framework for augmenting knowledge transfer across different graphs via multi-scale graph signal mixup. It consists of two major parts: Domain Unification and Trinity-signal Mixup, which give potential approaches to two challenges: \emph{C1. Graph Discrepancy} and \emph{ C2. Signal Heterogeneity} respectively. Extensive experimental results demonstrate the efficacy of our method for knowledge transfer across graphs. 




\printbibliography

\end{document}